\documentclass[runningheads]{llncs}
\usepackage[T1]{fontenc}
\usepackage{graphicx}
\usepackage{subcaption}
\usepackage{multicol}
\usepackage{multirow}
\usepackage{booktabs}
\usepackage{color}
\newcommand{\tb}[1]{\textcolor[rgb]{0.0,0.0,0.0}{#1}}    
\newcommand{\nate}[1]{\textcolor[rgb]{0.0,0.0,0.0}{#1}}    
\begin{document}
\title{\tb{Benchmarking NACTI Species Recognition\\in Long-Tailed Regimes}}
\titlerunning{Benchmarking NACTI in Long-Tailed Regimes}
%

\author{Zehua Liu\inst{1}\orcidID{0009-0004-5353-7028} \and
Tilo Burghardt\inst{1}\orcidID{0000-0002-8506-012X}}

%
%
\institute{School of Computer Science, University of Bristol, \\MVB Woodland Rd, BS8 1UB, Bristol, UK\\
\email{bw19062@bristol.ac.uk; tilo@cs.bris.ac.uk}\\
}
\maketitle              
\begin{abstract}

\nate{\tb{As with most ``in the wild'' collections of the natural world, the North America Camera Trap Images~(NACTI) dataset exhibits long-tailed class imbalance, with the largest class covering over~50\% of its~3.7M images. Building on the PyTorch Wildlife model, we systematically evaluate Long-Tail Recognition~(LTR) methodologies to benchmark species recognition performance, including specialised loss functions and LTR-sensitive regularisation. Our optimised configuration achieves state-of-the-art~99.40\% Top-1 accuracy on the NACTI test split, significantly outperforming standard baselines and previously reported top performances. To assess robustness under domain shifts~(e.g., night-time captures, occlusion, motion-blur), we extend our evaluation across three independent reduced-bias test sets~(including ENA-Detection, Caltech Camera Traps and Missouri Camera Traps). Across these out-of-distribution~(OOD) evaluations, our LTR-enhanced model consistently demonstrates substantially stronger generalisation capabilities compared to standard cross-entropy approaches. However, qualitative and quantitative analyses underline that current LTR optimisations cannot fully overcome representational bottlenecks, resulting in catastrophic predictive breakdown for rare `Tail' classes under severe domain shift. For maximum reproducibility, all dataset splits, key code, and network weights are published with this paper at \url{https://github.com/ZehuaLiuY/Species-Classification}.}}

\tb{}
\end{abstract}
\section{Introduction} 
\begin{figure*}[t]
    \centering
    \includegraphics[width=0.99\linewidth]{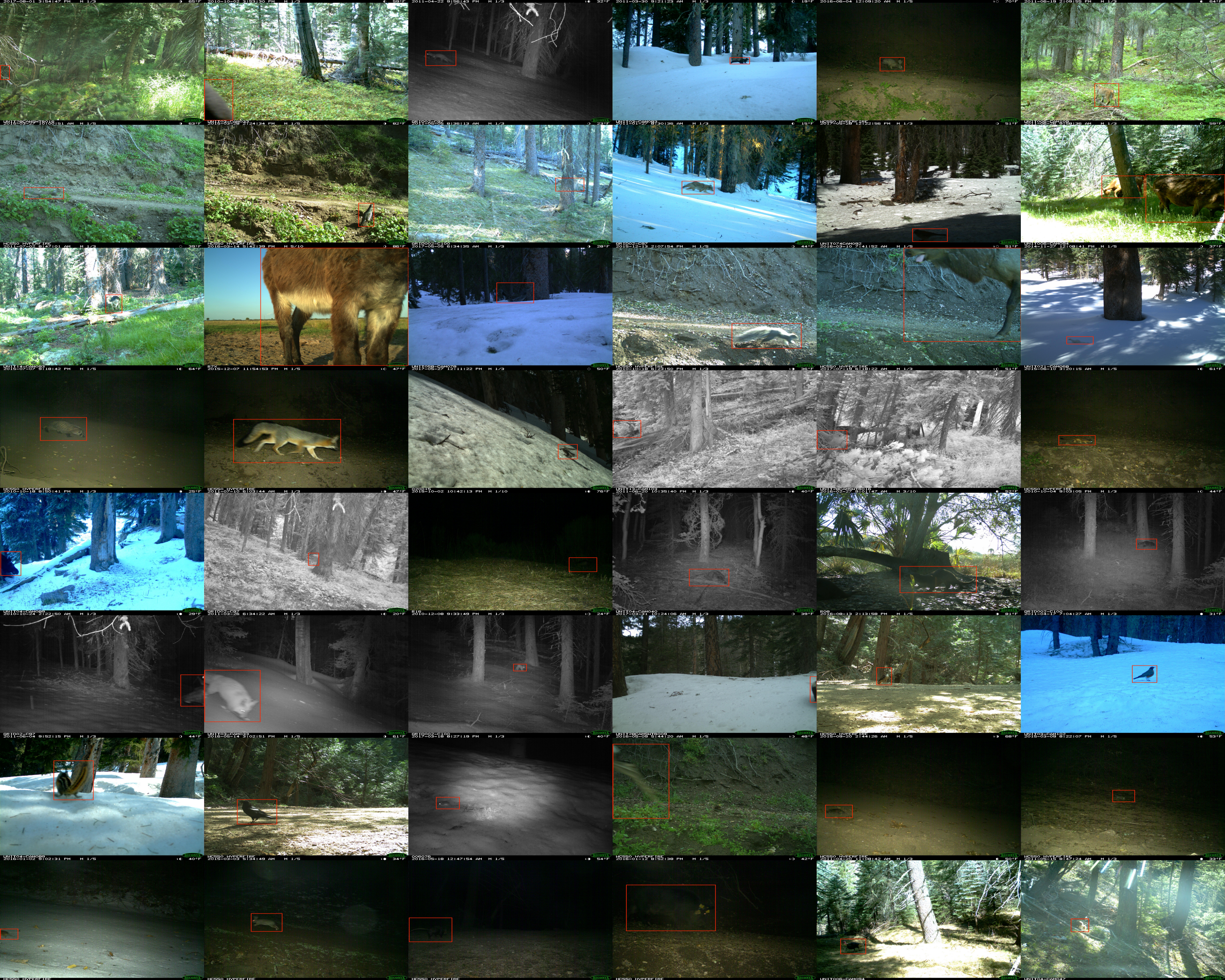}
    \captionof{figure}{\tb{\textbf{Representative Samples from the NACTI Dataset.}  Shown is one representative image per species class from the overall 48 classes in total, excluding empty and vehicle frames. Note challenging occlusions, background clutter, motion blur, infrared shots, and small animal areas. Red bounding boxes indicate detections using~\texttt{MegaDetectorV6}~\cite{beery2019efficient}, applied during the pre-processing stage. The NACTI dataset is long-tailed with over half of the samples belonging to a single species~(domestic cow). Following~\cite{perrett2023useheadimprovinglongtail}, 5 classes are categorised as the head, 37 form the ‘long tail’, and 6 occupy the few-shot end of the distribution.}}
    \label{fig:overview_graphic}
\end{figure*}

\paragraph{The Challenging NACTI Wildlife Dataset.}
\nate{\tb{The North American Camera Trap Images~(NACTI) dataset~\cite{NACTI2024} -- visually exemplified in Figure~\ref{fig:overview_graphic} contains 3.7M images across 48 animal classes where over half of its samples belong to a single species (domestic cow). Following Perrett et al.~\cite{perrett2023useheadimprovinglongtail}, 5 classes form the `Head', 37 form the `Long Tail', and 6 few-shot classes sit at the extreme end of the distribution. Beyond this stark statistical imbalance, the camera trap corpus suffers from a multitude of visual challenges encountered in the natural world: variations in lighting, motion blur, and inter-class species similarities all have the potential to degrade performance. Furthermore, the dataset's untrimmed images lack ground-truth bounding boxes or auxiliary per-instance metadata making fine-grained species ID challenging. Note, in this work we derive pseudo labels automatically using MegaDetector~\cite{beery2019efficient}~(see Section 3.2) while keeping the downstream task definition at image level, i.e. we treat NACTI as an image-level species classification problem rather than a joint detection-and-classification task.}}

\paragraph{Motivation and the Long-tail Challenge}
\nate{\tb{Accurate animal classification is essential for biodiversity conservation and ecological monitoring. With advances in computer vision, deep learning models, particularly Convolutional Neural Networks~(CNNs), have demonstrated remarkable capabilities in learning discriminative features from vast amounts of image data~\cite{CNN_features}. Tools like MegaDetector~\cite{beery2019efficient} and SpeciesNet~\cite{gadot2024crop} are trained on many millions of images and offer a strong foundation for general animal detection. However, they do not fully resolve the core issue of species-level classification under truly severe class imbalance. In real-world monitoring of local wildlife, datasets almost invariably exhibit the \textit{long-tail phenomenon} and show often \textit{severe class imbalance}. A substantial proportion of annotated data is typically comprised of common species, whereas only a small fraction represents rare, shy or endangered species. This skewed distribution affects machine learning models to often become biased towards majority classes, yielding high overall accuracy but catastrophically failing to identify rare specimen~\cite{Chen2014} - arguably the most valuable observation targets in conservation settings.}}

\begin{figure*}[h]
    \centering
    \includegraphics[width=0.99\linewidth,height=170pt]{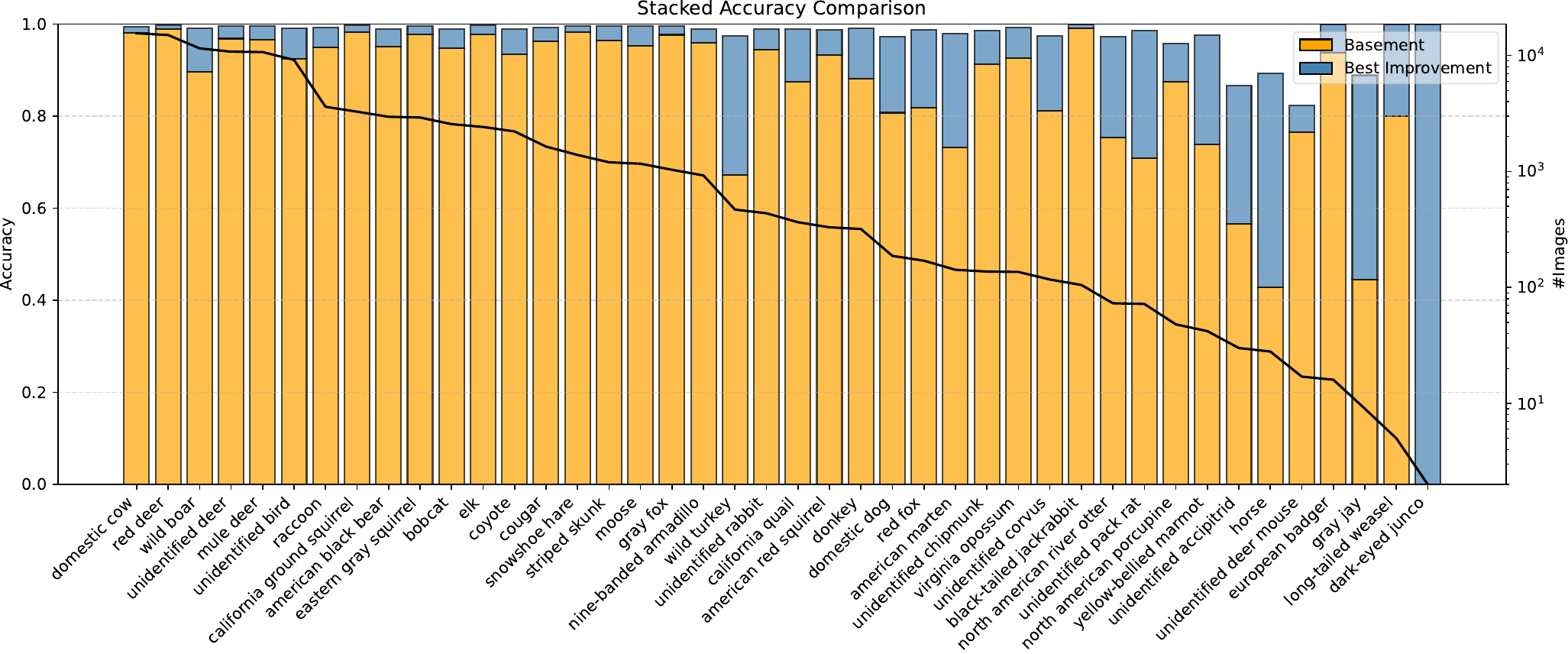}
    \captionof{figure}{{\tb{\textbf{LTR Performance Improvements across NACTI Species vs. Baseline.} Per-class Top-1 accuracies on the NACTI dataset comparing the baseline model (cross-entropy with Adam) against our best-performing configuration, which combines a Label-distribution-aware margin (LDAM) loss, an alternative AdamW optimiser, and a learning-rate scheduler. The results are reported on a 10\% random test split. The blue bars indicate classes with a positive improvement relative to the baseline.}}}\vspace{-5pt}
    \label{fig:baseline_vs_best}
\end{figure*}

\paragraph{Our Approach and Contributions.} 
\tb{Towards tackling this critical issue in computer vision for wildlife monitoring~\cite{tuia2022perspectives}, our work presents a systematic, loss-centric study focused on improving minority class detection performance in the large-scale, real-world NACTI setting as discussed in the associated conference paper~\cite{Liu2026}. We develop and present a scalable, multi-GPU pipeline to fine-tune state-of-the-art CNN architectures on the full NACTI corpus. Using this pipeline, we conduct a head-to-head comparison of Long-Tail Recognition (LTR) scheduling~\cite{alshammari2022longtailedrecognitionweightbalancing} and losses: focal loss~\cite{lin2018focallossdenseobject}, weighted cross-entropy~\cite{Cui2019ClassBalancedLoss}, and label-distribution-aware margin (LDAM) loss~\cite{cao2019learningimbalanceddatasetslabeldistributionaware}. Resulting from this analysis, Figure~\ref{fig:baseline_vs_best} summarises best model accuracy improvements over the baseline at class level. In addition to~\cite{Liu2026}, here we also benchmark results across out-of-distribution~(OOD) settings and discuss wider implications of species detection under unbalanced regimes and robustness under distributional shift. The line of work discussed in this paper has the following contributions, noting that the article is an extended version of the work presented at~\cite{Liu2026}:}

\nate{\tb{
\begin{enumerate}
  \item\textbf{Systematic NACTI LTR Evaluation.} In line with~\cite{Liu2026} we discuss the first head-to-head evaluation of LTR-sensitive scheduling in tandem with Focal, WCE, and LDAM losses for large-scale wildlife classification, analysing trade-offs between minority-class recall and overall accuracy.
  \item\textbf{Extensive Cross-Dataset OOD Robustness Analysis.} Moving the standard in-domain testing and extending~\cite{Liu2026} here, we rigorously evaluate model generalisation by constructing and testing against three independent reduced-bias datasets~(including ENA-Detection, Caltech Camera Traps, and Missouri Camera Traps). We demonstrate that while standard cross-entropy suffers under scheduled learning, the synergy between margin-aware LTR losses and schedulers yields substantial robustness gains guarding against some adverse effects of domain shifts.
  \item\textbf{Reproducible State-of-the-Art.} The line of work finally establishes and discuss here in full detail a reproducible 99.40\% Top-1 accuracy result on the NACTI test split~\cite{Liu2026}. We open-source our complete multi-GPU pipeline, data splits, and network weights to support transparent community research at~\cite{Liu2026}.
\end{enumerate}
}}

\section{Key Related Work}

\paragraph{Paper Context.} \nate{\tb{This article is an extended version of work originally presented at the 21st International Joint Conference on Computer Vision, Imaging and Computer Graphics Theory and Applications (VISAPP)~\cite{Liu2026}. While the conference paper established the baseline long-tailed recognition pipeline on the NACTI dataset, this manuscript significantly expands upon the previous research in several key dimensions. Specifically, this extended version introduces:}}

\begin{itemize}
    \item \tb{\textbf{Comprehensive OOD Evaluation:} We extend generalisation testing from a single biased evaluation to a robust, cross-dataset validation framework utilizing three independent reduced-bias test sets, notably incorporating the Missouri Camera Traps (MCT) and ENA-Detection datasets.}
    
    \item \tb{\textbf{Failure Mode Analysis:} We provide a substantially expanded quantitative and qualitative analysis of predictive collapse on tail classes~(e.g., \textit{horse} and~\textit{wild turkey}), demonstrating that representational bottlenecks persist across all experimental configurations.}
    
    \item \tb{\textbf{Generalisation Limitations:} We introduce new insights into the interplay between learning rate schedulers and LTR-specific loss functions under severe domain shift, concluding that purely optimisation-based methods are insufficient for bridging the representational gap in real-world wildlife monitoring.}
\end{itemize}

\subsection{Species Recognition in NACTI}

\paragraph{General ResNet-based Approaches.} \tb{Prior art on species recognition in NACTI goes back several years, but most published works lack reproducibility and published descriptions/code that would allow for a fair and detailed comparison of approaches to illuminate the effect of chosen architectures regarding the long-tail distribution issue. Tabak et al.~\cite{tabak2022cameratrapdetector}, for instance, trained a ResNet-50 architecture on the NACTI dataset (70/30 split) using SGD with momentum -- however, the paper does not explicitly specify the loss function used; though one would assume a softmax-cross-entropy setting was applied. They reported \(\approx\)86\% accuracy on the validation set. Performance dropped significantly for under-represented species (recall \(<\)70\%), highlighting persistent long-tailed bias in CamTrap data.}

\paragraph{Cross-Study MLWIC2 Project.} \tb{Tabak et al.~\cite{Tabak2020mlwic2} also trained MLWIC2, that is a ResNet-18 on $>3$M images from 18 North American studies (including NACTI), reporting 96.8\% Top-1 accuracy for the species model. To mitigate imbalance, the authors capped each class at 100{,}000 images. Despite these measures, rare species still showed weaker recall, indicating unresolved long-tail effects. For out-of-sample evaluations, reported Top-5 accuracy ranged from 65.2\% to 93.8\% depending on the dataset. We note, that the paper does neither specify optimiser nor loss function used.}

\paragraph{Active Learning.} \tb{Norouzzadeh et al.~\cite{Norouzzadeh2021DeepActiveLearning} combined a ResNet-50 with active learning techniques to iteratively select informative, unlabelled images for annotation. After 30{,}000 queries, the classifier reached 93.2\% accuracy on NACTI. In their pipeline, the embedding network is trained either using softmax-cross-entropy or with triplet loss settings. However, the exact evaluation split is not specified and the optimiser used for these experiments is not explicitly specified in the paper either (SGD is discussed only as general background), so we treat it as unspecified.}

\paragraph{Domain-aware DANAS Approach.} \tb{Jia et al.~\cite{jiadanas} introduced DANAS, a domain-aware neural architecture search tailored for camera-trap species classification. On the NACTI-a subset, DANAS showed that lightweight CNNs can match conventional performance baselines~(e.g., ResNet-18) within Top-1/Top-5 accuracy charts while using far fewer parameters and lower compute, enabling edge deployment at average accuracy \(\approx\)92.9\%. During the search phase, candidate networks were trained with AMSGrad~(learning rate = 0.005, batch size = 32) and guided by a customised, theoretically derived losses (Witch-of-Agnesi–based) that encourage high accuracy with few parameters. The optimiser used for the final training of the discovered DANAS networks is not explicitly stated, while baselines were stated and trained with Nesterov-SGD under a cosine loss regime~\cite{jiadanas}.}

\subsection{Regularisers and Losses for LTR}

\paragraph{Scheduling for LTR.} \tb{Addressing class imbalance, a central challenge in real-world data~\cite{zhang2023deeplongtailedlearningsurvey}, has led to various strategies. Recent study~\cite{alshammari2022longtailedrecognitionweightbalancing} have highlighted the importance of weight decay in mitigating classifier bias towards dominant classes, which is a key challenge in long-tailed recognition tasks. Alshammari et al.~\cite{alshammari2022longtailedrecognitionweightbalancing} showed that models trained without weight decay tend to assign disproportionately large classifier weights to frequent classes, resulting in poor generalisation for rare classes. By carefully tuning weight decay parameters, they significantly improved recognition performance on long-tailed datasets. Unlike conventional L1 or L2 regularisation, their method effectively balanced per-class classifier weights, thereby preventing common species from dominating learning. Furthermore, their study demonstrated that simple weight decay tuning outperformed many sophisticated LTR methods, suggesting that parameter regularisation should be of primary consideration when addressing class imbalance.}

\paragraph{Decoupled Learning Strategies.} \tb{A seminal work by Kang et al.~\cite{kang2020decouplingrepresentationclassifierlongtailed} proposed decoupling feature representation learning from classifier learning. They observed that standard instance-balanced sampling yields high-quality representations, while the classifier benefits from a class-balanced perspective, e.g. re-weighting or re-sampling. Their work also introduced LDAM-DRW, which combines the LDAM loss with a \textit{Deferred Re-weighting (DRW) schedule}, delaying the application of class-balancing until later in training. This concept strongly aligns with our findings, where combining the LDAM loss with a learning rate (LR) scheduler (ReduceLROnPlateau) yielded best results (see Section~\ref{ResLab} for details).}

\paragraph{Problem-specific LTR Loss Concepts.} \tb{To systematically address the severe class imbalance in the NACTI dataset, we implemented, compared, and evaluated three state-of-the-art loss functions that modify the standard cross-entropy in complementary ways. Our goal was to isolate the most effective strategy for improving performance on minority classes within a large-scale, real-world setting.}

\paragraph{Focal Loss.} \tb{Focal loss~\cite{lin2018focallossdenseobject} reduces the weight of easily classified samples, encouraging the model to focus more on hard-to-classify examples. This improves few-sample performance.}

\paragraph{Weighted Cross-Entropy~(WCE) Loss.} \tb{The WCE concept~\cite{Cui2019ClassBalancedLoss} addresses class imbalance by assigning greater weights to minority classes, thereby encouraging the model to focus more on under-represented categories during training and improving their recognition rates. However, this can lead to overfitting, causing the model to prioritise minority classes at the expense of overall performance.}

\paragraph{Label-Distribution-Aware Margin (LDAM) Loss.} \tb{LDAM~\cite{cao2019learningimbalanceddatasetslabeldistributionaware} addresses class imbalance by encouraging larger decision margins for minority classes. It modifies the standard cross-entropy loss by subtracting a class-dependent margin from the logit of the true class. This formulation gently penalises confident predictions for frequent classes while imposing stronger regularisation on minority classes. As a result, LDAM improves generalisation across the label distribution, especially for under-represented classes. It can also be combined with re-weighting or re-sampling strategies for further performance gains.}

\section{Data Preparation}
\subsection{Dataset Statistics}
\label{Section:Dataset}
\paragraph{The NACTI Dataset.} \tb{Published in the LILA Dataset collection~\cite{NACTI2024}, NACTI contains 3.7M camera-trap images from spanning 48 (sub)species. Figure~\ref{fig:overview_graphic} displays representative images from each species class, illustrating the wide variability in appearance, lighting, and environmental conditions. These samples exemplify common visual challenges encountered in camera trap imagery, including occlusions, background clutter, motion blur, and significant class imbalance. Such factors complicate the training process significantly particularly for under-represented classes where few images are available to capture species-distinctive features.}

\begin{figure*}[t]
  \centering
  \includegraphics[width=0.99\linewidth,height=180pt]{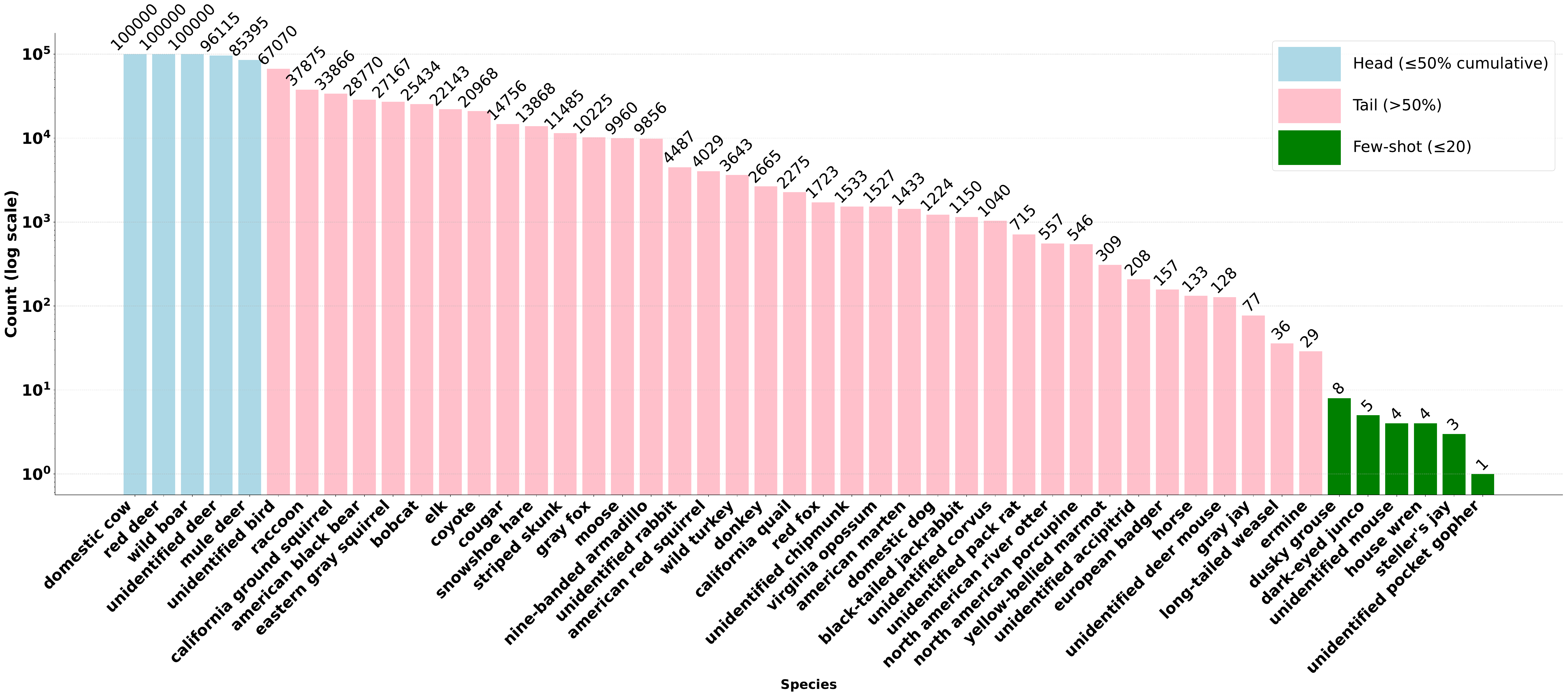}
  \caption{\tb{\textbf{NACTI Species Distribution after Sample Balancing.} Logarithmic visualisation of the cardinality of all 48 (sub)species classes of NACTI after capping classes at 100k. Colouration of the class distribution after~\cite{perrett2023useheadimprovinglongtail} breaking the distribution into `Head' (blue), `Tail' (pink), and `Few-shot' species.}}
  \label{fig:balanced_distribution}
\end{figure*}

\paragraph{Species Distribution Details.} \tb{The class distribution is extremely long-tailed; \textit{domestic cow} alone contributes 2,109,009 images (approx. $54\%$ of all samples). To reduce overfitting and speed up training, we cap each class at 100k images, following Tabak \textit{etal.}~\cite{Tabak2020mlwic2}. This yields 816,495 balanced samples (see Fig.~\ref{fig:balanced_distribution} for statistics) while preserving all 48 classes. The visualisation approach follows that of Perrett et al.~\cite{perrett2023useheadimprovinglongtail}, where the dataset is divided into three groups: `Head' classes (blue) contain $>50\%$ of total samples, `Tail' classes (pink) cover most species, while `Few-shot' classes (green) include those with $<20$ samples.}

\subsection{Data Preprocessing}
\tb{Our workflow converts the raw image NACTI archive into an analysis-ready dataset in three stages.}
\paragraph{(i) Automatic Animal Localisation.}  
 \tb{All images are first processed for entity detection with~\texttt{MegaDetectorV6} to obtain bounding boxes and confidence scores. The resulting JSON files are merged with the original metadata such that each record includes both species labels and coordinates for the detected animal region. In this study we crop the highest-confidence detection with a fixed margin and use that crop as the input that is resized.}

\paragraph{(ii) Class-Balanced Resampling.}  
\tb{Following Tabak~\emph{et al.}, we cap every category at 100,000 samples, discard empty/vehicle frames, and retain all 48 animal classes. This produces 816,495 images with a more balanced distribution profile shown in Figure~\ref{fig:balanced_distribution}.}

\paragraph{(iii) Split Strategy and Reduced-Bias Evaluation.}  
\tb{The balanced pool is split \textit{80/10/10} (train/validation/test) with a fixed seed, acknowledging that visually similar burst information may still be spread across subsets in this setting. To gauge towards bias reduction and better generalisation we also construct and publish a \emph{reduced-bias test set} using the ENA Dataset~\cite{ena2024}, filtering to the nine classes shared with NACTI and removing any temporal/spatial overlap. All nine fall in the distribution `Tail’, providing a stringent domain-shift benchmark.}


\paragraph{Final Pre-processing Steps.} \tb{Images are resized to $256\!\times\!256$ px as per Tabak~\emph{et al.}, converted to tensors, and normalised with \(\mu=[0.485,0.456,0.406]\), \(\sigma=[0.229,0.224,0.225]\) to stabilise optimisation. Each training sample is finally stored as an information tuple~\texttt{(image, \{bbox, label, conf\})}.}

\paragraph{Method Overview.} \tb{We adopt the pre-trained ResNet-50 AI4GAmazonRainforest framework as our feature extractor backbone and fine-tune a 48-way SoftMax classifier head for NACTI species prediction. We systematically compare standard cross-entropy results against three long-tail-aware alternatives -- focal loss, class-balanced weighted cross-entropy (WCE), and label-distribution-aware margin (LDAM) loss—in their standard formulations from the literature. {The resulting models are evaluated on the balanced NACTI test split and the reduced-bias test sets, isolating loss and scheduling choices for in- and out-of-domain performance.}}

\begin{figure*}[t]
        \centering
        \includegraphics[width=0.8\linewidth]{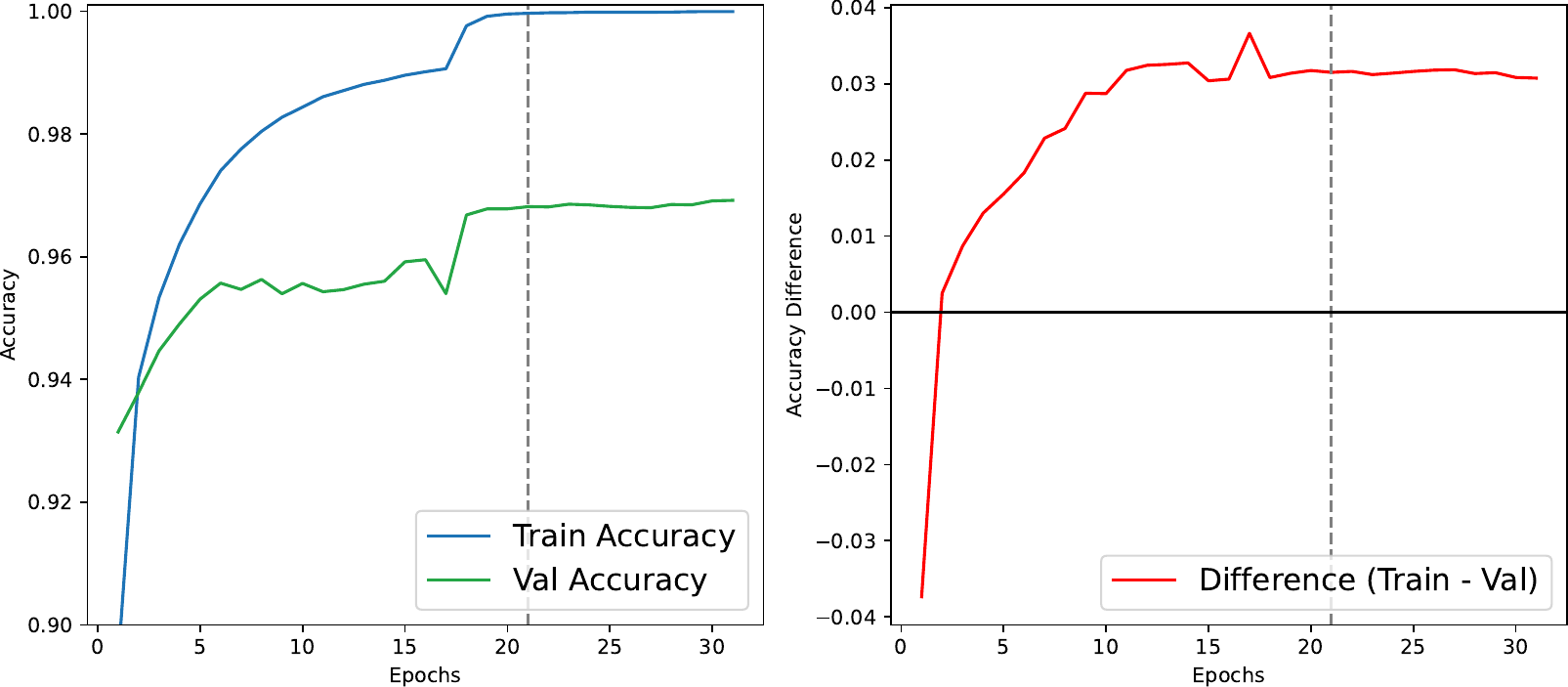}
    \caption{\tb{\textbf{Training Dynamics - Accuracy.} Accuracy development during model training, the left plot shows train and validation while the right plots show the train-validation gap. Note, the dashed lines marking early-stopping point determined based on val performance.}}
    \label{fig:training_dynamicsa}
\end{figure*}

\section{Experiments}
\subsection{Baseline Model}
\paragraph{Hardware Specification.} \tb{All training is conducted utilising 11 NVIDIA Tesla P100 GPUs~(Pascal architecture) to efficiently distribute computation across multiple devices and reduce training time. All inference was performed locally on an NVIDIA GeForce RTX 3090 GPU~(Ampere architecture).}

\begin{figure}[t]
 \centering
        \includegraphics[width=0.8\linewidth]{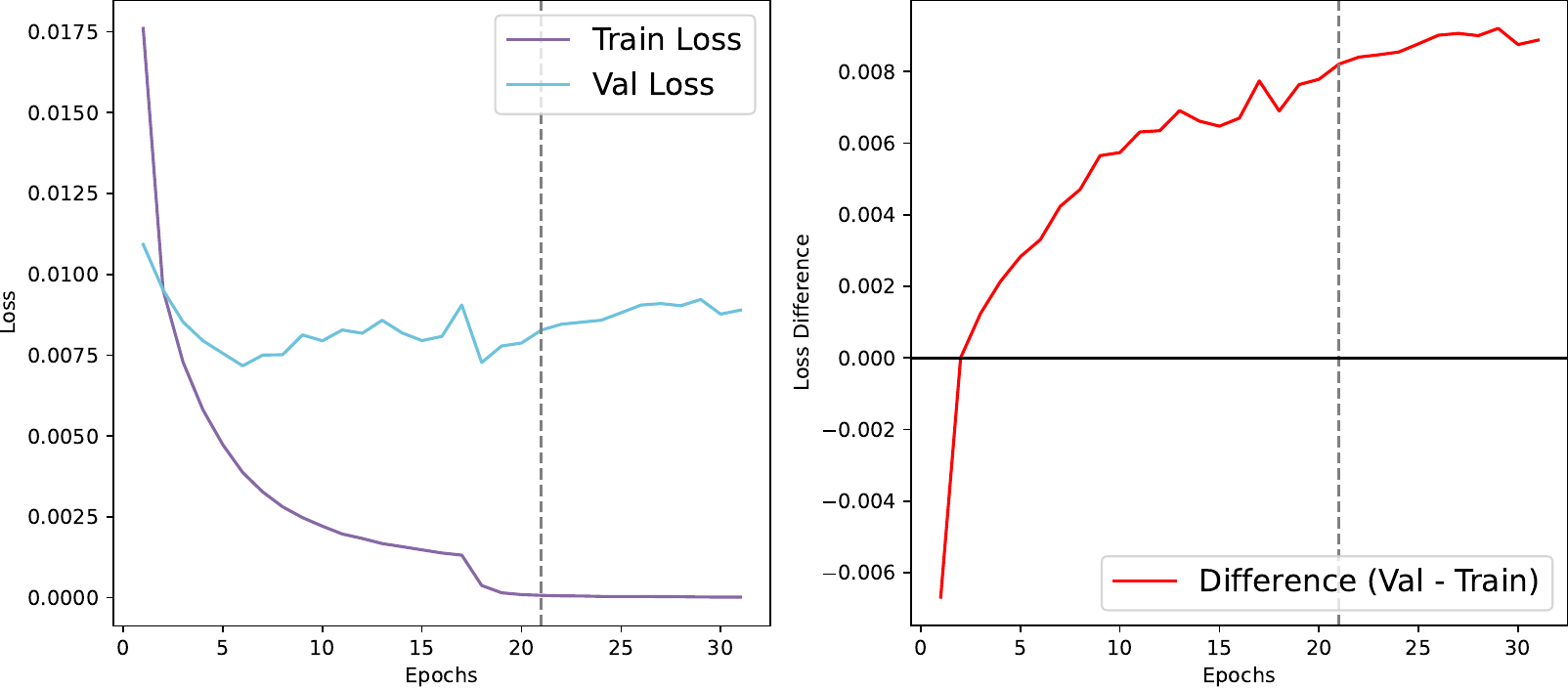}
\vspace{-10pt}
    \caption{\tb{\textbf{Training Dynamics - Loss.} Loss development during model training, the left plot shows train and validation while the right plots show the train-validation gap. Note the dashed lines marking early-stopping point, where stopping is determined on val accuracy not loss (see Fig~\ref{fig:training_dynamicsa} for this).}}
    \label{fig:training_dynamics}
\end{figure}

\paragraph{Full Network Training Details.} \tb{All experiments fine-tune the ResNet-50~\cite{he2016deep} backbone of \texttt{AI4GAmazonRainforest}~\cite{hernandez2024pytorchwildlife}  with its standard pre-trained weights, using AdamW~\cite{Loshchilov2017DecoupledWD} (weight-decay $10^{-2}$, initial learning rate $10^{-4}$) and importantly a \texttt{ReduceLROnPlateau} scheduler that lowers the learning rate by factor 0.1 when validation metrics stagnate. To avoid premature termination or overfitting caused by fluctuations in validation performance we employ early stopping (see Figure~\ref{fig:training_dynamics}) with patience 10 and check-pointing, dynamically selecting the best model based on validation recall~(see published code).}

\subsection{Experimental Results}\label{ResLab}
\paragraph{Scheduler Impact on NACTI Results.} \tb{Table~\ref{tab:configuration_comparison_biased_tail} compares various combinations of LTR-sensitive learning rate scheduling and LTR loss functions+optimisers against baselines, evaluated in terms of both overall accuracy and performance on the `Tail' classes. Introducing a learning rate scheduler from PyTorch~\cite{pytorch} resulted in substantial performance gains across all configurations, with overall accuracy consistently exceeding 99\% on the biased test set. The overall best result was achieved by combining LDAM loss with the AdamW optimiser and the \texttt{ReduceLROnPlateau} scheduler, yielding an accuracy of 99.40\%—an improvement of approximately 2.66 percentage points over the same configuration without a scheduler (96.74\%).}
\begin{table}[t]\vspace{8pt}
    \caption{\tb{\textbf{NACTI Test Accuracy (LTR Methods vs. Baselines)} Comparison of (Overall/`Tail') accuracy on the biased NACTI test set across all configurations. Note that `Tail' performance stats are saturated with the top accuracy of 99.22\% is shared by three configurations.}}\vspace{5pt}
    \centering
    \small
        \setlength{\tabcolsep}{3pt}
    \begin{tabular}{|l|c|c|}
    \hline
    \textbf{ LOSS + OPTIMISER } & \textbf{BASELINES} & \textbf{LTR LR}\\
        \textbf{ CONFIGURATION} & { (NO SCHEDULER) } & \textbf{ SCHEDULER} \\
    \hline
     \textbf{\textit{BASELINES}}   &  &  \\
    Cross-Entropy + Adam   & 95.51\% / 94.30\% & \emph{- not tested -} \\
    Cross-Entropy + AdamW  & 96.77\% / 95.64\% & 99.34\% / \textbf{99.22\%} \\
        \hline
    \textbf{\textit{LTR LOSSES}}   &  &  \\
    Focal Loss + AdamW  & 96.65\% / 95.96\% & 99.31\% / 99.20\%  \\
    WCE + AdamW & 96.69\% / 96.16\% & 99.35\% / \textbf{99.22\%} \\
    LDAM Loss + AdamW & 96.71\% / 96.17\% & \textbf{99.40\%} / \textbf{99.22\%} \\
    \hline
    \end{tabular}
    \label{tab:configuration_comparison_biased_tail}
\end{table}

\paragraph{NACTI Tail Results.} \tb{A similar trend was observed for the `Tail' classes: accuracy increased from 96.17\% to 99.22\%, representing a relative improvement of 3.05\%. Although all configurations achieved comparable performance on the `Tail' classes following scheduler integration, the LDAM + AdamW combination attained the highest overall accuracy and is henceforth referred to as \textit{best (LTR) configuration}.}


\paragraph{Cross-Dataset OOD Evaluation Results.} \tb{To comprehensively assess generalisation performance under severe distribution shift, all configurations are additionally evaluated across three independent out-of-distribution benchmarks. Specifically, we constructed three distinct Reduced-Bias Test Sets derived from the ENA-Detection, Missouri Camera Traps (MCT), and Caltech Camera Traps (CCT) datasets, respectively. Table~\ref{tab:configuration_comparison_unbiased} summarises this task set, showing expected and substantial accuracy drops across the board relative to the biased in-domain evaluation setup.}

\begin{table}[bp]
    \caption{\nate{\tb{\textbf{Cross-Dataset OOD Generalisation Accuracy (\%).} Accuracy on ENA, MCT, and CCT datasets. Exact mappings of class labels are given in full in the code base. The combination of LDAM and LR scheduling yields the highest robustness and improve benchmarks slightly compared to stronger collapse in baseline settings.}}}\vspace{5pt}
    \centering
    \small
    \setlength{\tabcolsep}{3pt}
    \begin{tabular}{|l|cc|cc|cc|}
        \hline
        \multirow{2}{*}{\textbf{Configuration}} & \multicolumn{2}{c|}{\textbf{ENA}} & \multicolumn{2}{c|}{\textbf{MCT}} & \multicolumn{2}{c|}{\textbf{CCT}} \\
        \cline{2-7}
        & \textbf{Base.} & \textbf{Sched.} & \textbf{Base.} & \textbf{Sched.} & \textbf{Base.} & \textbf{Sched.} \\
        \hline
        CE (Adam) & 39.48 & - & 32.78 & - & 55.56 & - \\
        CE (AdamW) & 38.27 & 48.27 & 33.50 & 31.91 & 52.90 & 62.09 \\
        \hline
        Focal Loss (AdamW) & 37.62 & 49.78 & 15.49 & 22.45 & 51.13 & 62.99 \\
        WCE (AdamW) & \textbf{45.03} & 51.20 & 24.53 & 32.71 & \textbf{64.45} & 63.66 \\
        LDAM Loss (AdamW) & 38.77 & \textbf{52.55} & \textbf{38.95} & \textbf{35.71} & 52.55 & \textbf{63.75} \\
        \hline
    \end{tabular}
    \label{tab:configuration_comparison_unbiased}
\end{table}
\paragraph{Nuanced Impact of LTR Losses and Scheduling.} \tb{While performance degrades in all OOD settings, the integration of LTR-specific loss functions and learning rate schedulers reveals a profound and nuanced trade-off depending on the class frequency distribution of the target domain. For benchmarks predominantly composed of `Tail' classes (ENA and CCT), the synergy between margin-aware LTR losses and scheduled learning generally yields strong robustness. For instance, the LDAM loss combined with a scheduler achieves 52.55\% on ENA and 63.75\% on CCT, securing substantial gains over standard cross-entropy baselines. However, a starkly different pattern emerges on the MCT dataset, where both standard and LTR configurations perform significantly worse overall, and the application of a scheduler actively degrades performance for the LDAM loss (dropping from 38.95\% to 35.71\%). This divergence indicates that LTR optimisations do not uniformly improve cross-domain generalisation, pointing to a class-specific representational bottleneck.}

\section{Discussion}
\subsection{Generalisation Considerations}
\paragraph{Testing Bias and Performance Saturation.}  \tb{As discussed in Section~\ref{Section:Dataset}, the standard NACTI Test Set -- using the standard random selection process for testing NACTI so far -- contains numerous visually similar frames originating from camera-trap burst capture, making it especially vulnerable to biased testing. As a result, test accuracy fails to reflect generalisability and leading to saturated 99\%+ performances. However, the model's ability to generalise is shown on the reduced-bias test set where different LTR loss functions can offer minor accuracy improvements as seen in Table~\ref{tab:configuration_comparison_unbiased}. The dramatic performance gap between biased and reduced-bias test sets underscores the need for model-external OOD evaluation particularly for wildlife data like NACTI where significant context changes are part of environmental variability.}


\paragraph{Per-Class Generalisation Analysis.}
\tb{To understand the generalisation behaviour of the best-performing model further on the Reduced-Bias Dataset, per-class accuracies for the 9 classes in this dataset are reported in Table~\ref{tab:nonzero_accuracy_with_horse}. Note that all listed species belong to the `Tail' of the training distribution. While the model demonstrates relatively strong performance on classes such as \textit{coyote} and \textit{american black bear}, accuracy varies considerably across species. Notably, some classes like \textit{wild turkey} and \textit{horse} receive 0\% accuracy. These results underscore the difficulty of long-tailed generalisation under distribution shift. These findings suggest that while optimisation strategies such as LTR loss functions and learning rate schedules can improve overall accuracy, operation on rare classes may underperform catastrophically under distribution shift. Persistently low or zero accuracy on certain species across all configurations indicates that representational limitations play a more significant role unrecoverable by means of current learning alterations, potentially requiring a data-centric approach.}

\subsection{ENA Detection dataset Failure Mode Analysis}
\label{sec:consistent_failures}
\begin{table}[t]
    \caption{\tb{\textbf{Per-class Accuracy and Prevalence.} Results are for the ENA-Detection under our best configuration.}}\vspace{5pt}
    \centering
    \small
    \begin{tabular}{|p{3.4cm} | r | r|}
    \hline
    \textbf{CLASS} & \textbf{ACCURACY (\%)} & \textbf{PREVALENCE}\\
    \hline
    virginia opossum & 65.79\% & 725  \\
    bobcat & 61.26\% & 333 \\
    american black bear & 77.05\% & 475  \\
    eastern gray squirrel & 45.51\% & 312 \\
    wild turkey & 0.00\% & 427  \\
    striped skunk & 21.55\% & 297 \\
    red fox & 54.72\% & 413 \\
    horse & 0.00\% & 63 \\
    coyote & 87.79\% & 344  \\
    \hline
    \end{tabular}%
    \label{tab:nonzero_accuracy_with_horse}
\end{table}
\begin{figure}[t]
  \centering
    \includegraphics[width=0.99\linewidth]{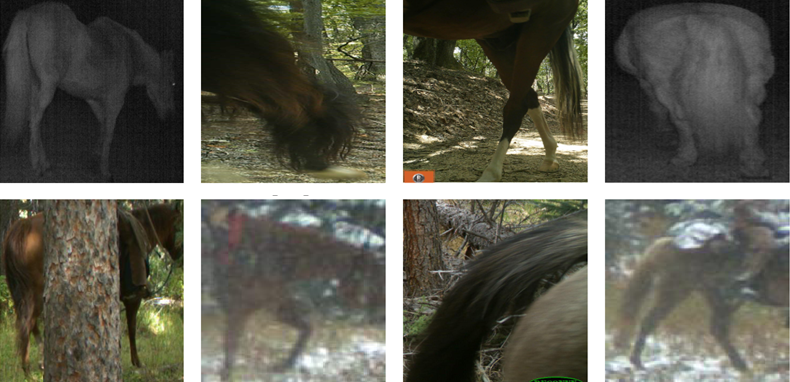}
    \label{fig:horse_examples_sub}
  \caption{\tb{\textbf{Qualitative Examples regarding Domain Shift for Horse Class.} Train-test contrast of horse class with little image evidence in training where catastrophic performance breakdown was observed under distribution shift. Top row shows examples from bias-reduced test set whilst bottom row shows samples from the training set. Training images are often blurry or occlude major animal parts, whereas reduced-bias test images involve night scenes, partial views etc.}}
  \label{fig:domain_shift_examples}
\end{figure}

\paragraph{Predictive Collapse for Tail Classes.} \tb{Two classes -- \textit{horse} (\(n=63\)) and \textit{wild turkey} (\(n=148\)) --  consistently rank among the lowest-performing classes in the biased NACTI test set across \emph{all} configurations and accuracy for both classes collapses to approximate 0\% in the Reduced-Bias Test Set.  Tables~\ref{tab:horse_preds} exemplifies this collapse quantitatively. The model struggles to learn fine-grained distinctions between semantically or morphologically similar species when the training data does not contain sufficient examples of classes that actually contain species-determining features. This poses an ill-posed task of recognition of species who have little to no definition during the training phase in the first place. The consistent yet unsurprising misclassification into frequent categories confirms a strong prediction bias toward `Head' classes, reflecting the model’s defaulting to dominating features learned from majority categories.}

\begin{figure}[t]
  \centering
    \includegraphics[width=0.99\linewidth]{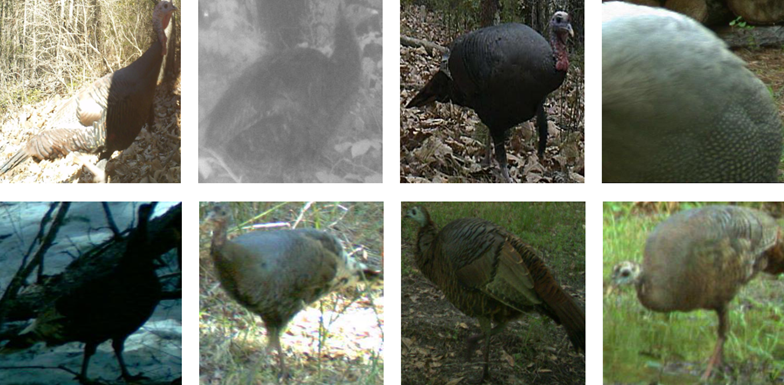}
    \label{fig:horse_examples_sub}
  \caption{\tb{\textbf{Qualitative Examples regarding Domain Shift for Turkey Class.} Train-test contrast of turkey class with, like horses, a small image corpus in training in the first place. Top row shows examples from bias-reduced test set whilst bottom row shows samples from the training set. Test images again involve night scenes, extreme shadows, and partial occlusions.}}
  \label{fig:domain_shift_examples2}
\end{figure}

\paragraph{Qualitative Evidence.} \tb{Figures~\ref{fig:domain_shift_examples} and~\ref{fig:domain_shift_examples2} juxtapose representative images for challenging tail classes (e.g., \textit{horse} and \textit{wild turkey}) from the NACTI training set and the Reduced-Bias Test Set, illustrating severe visual domain discrepancies. For the \textit{horse} class, training images are often blurry or partially occluded, making them suboptimal for learning robust features. In contrast, the reduced-bias test instances frequently introduce night-time infrared captures, unusual poses, or display only a partial view of the animal. This forces the model to rely on visual fragments, leading to prediction errors where horses are misclassified as visually similar, more frequent classes such as \textit{red deer} or \textit{domestic cow}.}

\tb{A similarly pronounced domain gap is observed for the \textit{wild turkey} class. While its training samples are typically high-resolution and captured in clear natural outdoor environments, test set images frequently suffer from heavy partial occlusion, cluttered backgrounds, motion blur, and poor lighting. Without full-body visibility or clear contextual cues, wild turkeys in the test set may appear visually similar to other species in terms of body shape or feather texture. This visual ambiguity prompts the model to systematically default to visually proximate categories like \textit{unidentified bird} or dominant head classes like \textit{wild boar}. These visual discrepancies form a significant representational gap, yielding unreliable feature representations and poorly defined decision boundaries that severely restrict the model’s capacity for effective generalisation. Ultimately, two interacting factors contribute to this total detection collapse:}

\begin{enumerate}
    \item \tb{\textbf{Class-Imbalance Bias.} Misclassifications of severely under-represented classes during training occur due to the network defaulting to dominant visual cues learned from the majority classes, yielding skewed posteriors that rely on class distribution priors rather than class-indicative image data.}
    \item \tb{\textbf{Domain Shift.} Compared with training images, our Reduced-Bias Test shots exhibit different lighting, more partial occlusion, night scenes and stronger motion blur overall.  These discrepancies widen the representational gap and can lead to total detection collapse~(see Table~\ref{tab:horse_preds} for \emph{'horse'} class and Table~\ref{tab:turkey_preds} for \emph{'wild turkey'} class).}
\end{enumerate}

\paragraph{Partial View Considerations.} \tb{For many tail classes including \textit{horse}, the network almost invariably predicts dominant classes, for \textit{horse} mainly \textit{red deer} or \textit{domestic cow} whenever a full body is not visible. This not only reflects the model’s bias towards the most frequent `Head’ classes under uncertainty, but points particularly to a problem of `unseen' views. This observation points in the direction of future work to analyse data augmentation and the simulation of partial view detection and related generative data strategies to compliment LTR scheduling.}

\begin{table}[t]\vspace{8pt}
  \centering
  \caption{\tb{\textbf{Collapse in OOD Classification for Class~\textit{`horse'}.} Distribution of class recognitions (all false) of best performing LTR model in the Reduced-Bias Test Set confirms catastrophic collapse of classification with the  classifier defaulting to head classes even under maximal use of LTR adjustments.}}\vspace{5pt}
  \label{tab:horse_preds}
  \begin{tabular}{|l|r|r|}
    \hline
    \textbf{ PREDICTED CLS }       & \textbf{ COUNT } & \textbf{ PERCENT } (\%) \\ 
    \hline
    domestic cow           & 28 & 44.4 \\
    red deer               & 19 & 30.2 \\
    coyote                 &  6 &  9.5 \\
    cougar                 &  4 &  6.3 \\
    american black bear    &  3 &  4.8 \\
    unidentified deer      &  1 &  1.6 \\
    mule deer              &  1 &  1.6 \\
    unidentified bird      &  1 &  1.6 \\
    \hline
  \end{tabular}
\end{table}\vspace{5pt}

\subsection{MCT Dataset Failure Mode Analysis}
\paragraph{The LTR Trade-off on Head Classes under Domain Shift.}\tb{
To understand the catastrophic breakdown and the negative impact of scheduling on the MCT dataset, we must analyse its species composition through the lens of LTR loss mechanisms. Unlike ENA and CCT which consist entirely of `Tail' species, the MCT dataset prominently features major `Head' classes from the NACTI training distribution, specifically \textit{red deer} and \textit{wild boar}. LTR optimisation strategies, particularly LDAM and weighted configurations, are explicitly designed to penalise frequent classes by enforcing stricter margins or down-weighting their gradients, thereby forcing the model to allocate more representational capacity to rare species. While this ``suppress the head to boost the tail'' logic succeeds in-domain, our MCT results expose its fragility under domain shift. Because the decision boundaries for `Head' classes are artificially restricted during LTR training, the model's feature representation for these species becomes highly sensitive. When faced with degraded, out-of-distribution images of \textit{red deer} or \textit{wild boar} in the MCT dataset, these shifted features easily fall outside their restricted margins. Consequently, further applying a learning rate scheduler—which aggressively cements these LTR decision boundaries during late-stage training—exacerbates the misclassification of OOD `Head' classes, explaining the counter-intuitive performance drop (e.g., LDAM dropping to 35.71\%).}

\begin{table}[t]\vspace{8pt}
  \centering
  \caption{\nate{\textbf{Collapse in OOD Classification for Class~\textit{`wild turkey'}.} Distribution of class recognitions (all false) of best performing LTR model in the Reduced-Bias Test Set confirms catastrophic collapse of classification with the  classifier defaulting to head classes even under maximal use of LTR adjustments.}}\vspace{5pt}
  \label{tab:turkey_preds}
  \begin{tabular}{|l|r|r|}
    \hline
    \textbf{ PREDICTED CLS }       & \textbf{ COUNT } & \textbf{ PERCENT } (\%) \\ 
    \hline
    unidentified bird                     & 193 & 45.2 \\
    wild boar                             & 113 & 26.5 \\
    nine-banded armadillo                 &  26 &  6.1 \\
    red deer                              &  19 &  4.4 \\
    mule deer                             &  17 &  4.0 \\
    unidentified corvus                   &  16 &  3.7 \\
    domestic cow                          &  15 &  3.5 \\
    domestic dog                          &  9 &  2.1 \\
    virginia opossum                      & 8 & 1.9 \\
    american black bear                   & 6 & 1.4 \\
    unidentified chipmunk                 & 2 &  $\approx$ 0.0 \\
    coyote                                &  1 &  $\approx$ 0.0 \\
    horse                                 &  1 &  $\approx$ 0.0 \\
    california quail                      &  1 &  $\approx$ 0.0 \\
    \hline
  \end{tabular}
\end{table}\vspace{5pt}

\paragraph{Misclassification Patterns of OOD Head Classes.} \tb{
To quantitatively substantiate this trade-off, Table~\ref{tab:mct_head_class_drop} isolates the performance of the two dominant `Head' classes (\textit{red deer} and \textit{wild boar}) within the MCT dataset. The data reveals a stark performance degradation or total predictive collapse when the learning rate scheduler is introduced to the LDAM loss. For instance, the accuracy for \textit{red deer} explicitly drops from 71.55\% to 65.48\%, with the model increasingly defaulting to visually proximate but incorrect overarching classes such as \textit{domestic cow}. Similarly, \textit{wild boar} suffers from near-total detection failure (0.05\% accuracy), being systematically misclassified into the other prominent majority class, \textit{red deer}. This targeted predictive collapse directly correlates with the overall performance drop observed in Table~\ref{tab:configuration_comparison_unbiased}. It provides empirical evidence that while LTR scheduling refines decision boundaries for rare species in-domain, it impacts the generalisation potential for the feature subspace of majority classes. When these majority classes undergo severe environmental shifts in natural OOD scenarios as in wildlife data, they are often pushed outside their \#margins towards alternative dominant priors leading to degradation of OOD performance even in head classes. This highlights that current training datasets like NACTI may contain rich species information, but the naturally possible variation of species' appearance is not captured in them fully with significant need for further research.}

\begin{table}[t]
    \caption{\tb{\textbf{Predictive Collapse of `Head' Classes on the MCT Dataset.} Comparison of LDAM performance with and without a learning rate scheduler for the two dominant `Head' species. The scheduler exacerbates accuracy degradation for OOD majority classes, forcing reliance on alternative dominant priors.}}\vspace{5pt}
    \centering
    \small
    \begin{tabular}{l c c c}
        \toprule
        \multirow{2}{*}{\textbf{Head Class}} & \multicolumn{2}{c}{\textbf{Accuracy (\%)}} & \multirow{2}{*}{\begin{tabular}[c]{@{}c@{}}\textbf{Top Misclassified} \\ \textbf{Class (Sched.)}\end{tabular}} \\
        \cmidrule(lr){2-3}
        & \textbf{LDAM (Base)} & \textbf{LDAM + Sched.} & \\
        \midrule
        red deer & 71.55\% & 65.48\% & \textit{domestic cow} \\
        wild boar & 0.00\% & 0.05\% & \textit{red deer} \\
        \bottomrule
    \end{tabular}
    \label{tab:mct_head_class_drop}
\end{table}

\newpage
\section{Conclusions}
\label{sec:conclusion}

\paragraph{Reproducible NACTI SOTA Performance.} \tb{In this line of work we delivered a reproducible improvement in long-tailed species recognition for the NACTI dataset. To make these capabilities accessible to the community, we publish the end-to-end pipeline that processes raw NACTI frames and fine-tunes a ResNet-50-based architecture with LTR losses under Adam/AdamW optimisers. We confirm experimentally that a scheduled LDAM/AdamW configuration will yield overall 99.4\% Top-1 accuracy, setting a new reference point for NACTI wildlife species recognition under long-tailed conditions. We publish full code, weights, and complete data split information for reproducibility.}

\paragraph{OOD Analysis and Recommendation.} \tb{We also provided a significantly extended domain-shift evaluation across three independent benchmarks (ENA, MCT, and CCT), exposing a critical representational trade-off in current long-tailed recognition methods. We demonstrated that while combining margin-aware LDAM losses with learning rate scheduling yields optimal generalisation for `Tail'-heavy OOD sets (e.g., ENA and CCT), this identical configuration may aggressively degrade performance on OOD `Head' classes (MCT) in different settings. This shows an unfavourable interaction between domain shifts in natural settings and LTR adjustments to counteract long-tail distributions. Thus, given these insights in context with previous work~\cite{alshammari2022longtailedrecognitionweightbalancing}, the recommendation of LDAM-scheduler setups as a starting point for boosting rare species detection must be given with the caveat that purely optimisation-based LTR methods remain insufficient for holistic cross-domain robustness across the entire data distribution. By establishing a transparent benchmark, detailing these failure modes, and releasing reproducible SOTA pipelines, we hope to encourage the development of robust, data-centric, and transparent solutions in the evolving field of computer vision for wildlife monitoring.}

\end{document}